\documentclass{article}

\usepackage[final]{neurips_2023}

\usepackage[utf8]{inputenc} 
\usepackage[T1]{fontenc}    
\usepackage{hyperref}       
\usepackage{url}            
\usepackage{booktabs}       
\usepackage{amsfonts}       
\usepackage{nicefrac}       
\usepackage{microtype}      
\usepackage{xcolor}         
\usepackage[pdftex]{graphicx} 
\usepackage{colortbl}       
\usepackage{algorithm}      
\usepackage{amsmath}        
\usepackage{mathtools}
\usepackage[font=footnotesize]{caption}
\usepackage{wrapfig}
\usepackage{bbm}

\title{Estimating Uncertainty in Multimodal Foundation Models using Public Internet Data}

\author{
  Shiladitya Dutta*\\
  UC Berkeley\\
  \And
  Hongbo Wei*\\
  UC Berkeley\\
  \And
  Lars van der Laan\\
  University of Washington\\
  \And
  Ahmed M. Alaa\\
  UC Berkeley and UCSF\\
}

\begin{document}

\maketitle

\begin{abstract}
Foundation models are trained on vast amounts of data at scale using self-supervised learning, enabling adaptation to a wide range of downstream tasks. At test time, these models exhibit {\it zero-shot} capabilities through which they can classify previously unseen (user-specified) categories. In this paper, we address~the~problem~of quantifying uncertainty in these zero-shot predictions. We~propose~a~heuristic~approach for uncertainty estimation in zero-shot settings using {\it conformal prediction} with web data. Given a set of classes at test time, we~conduct~zero-shot~classification with CLIP-style models using a prompt template, e.g., {\it ``an~image~of~a~<category>''}, and use the same template as a search query to source calibration data~from~the~open web. Given a web-based calibration set, we apply conformal~prediction~with~a~novel conformity score that accounts for potential errors in retrieved web data. We evaluate the utility of our proposed method in Biomedical~foundation~models;~our~preliminary results show that web-based conformal prediction sets achieve the target coverage with satisfactory efficiency on a variety of biomedical datasets.
\end{abstract}
\vspace{-0.1in}

\,\,\,\,\,\,\,\,\,\,\,\,\,\,\,\,\,\,\,\,\,{\bf Code:} \href{https://github.com/AlaaLab/WebCP}{https://github.com/AlaaLab/WebCP}, * Equal Contribution.

\vspace{0.25in}
\begin{figure}[h] 
  \centering
  \includegraphics[width=5.5in]{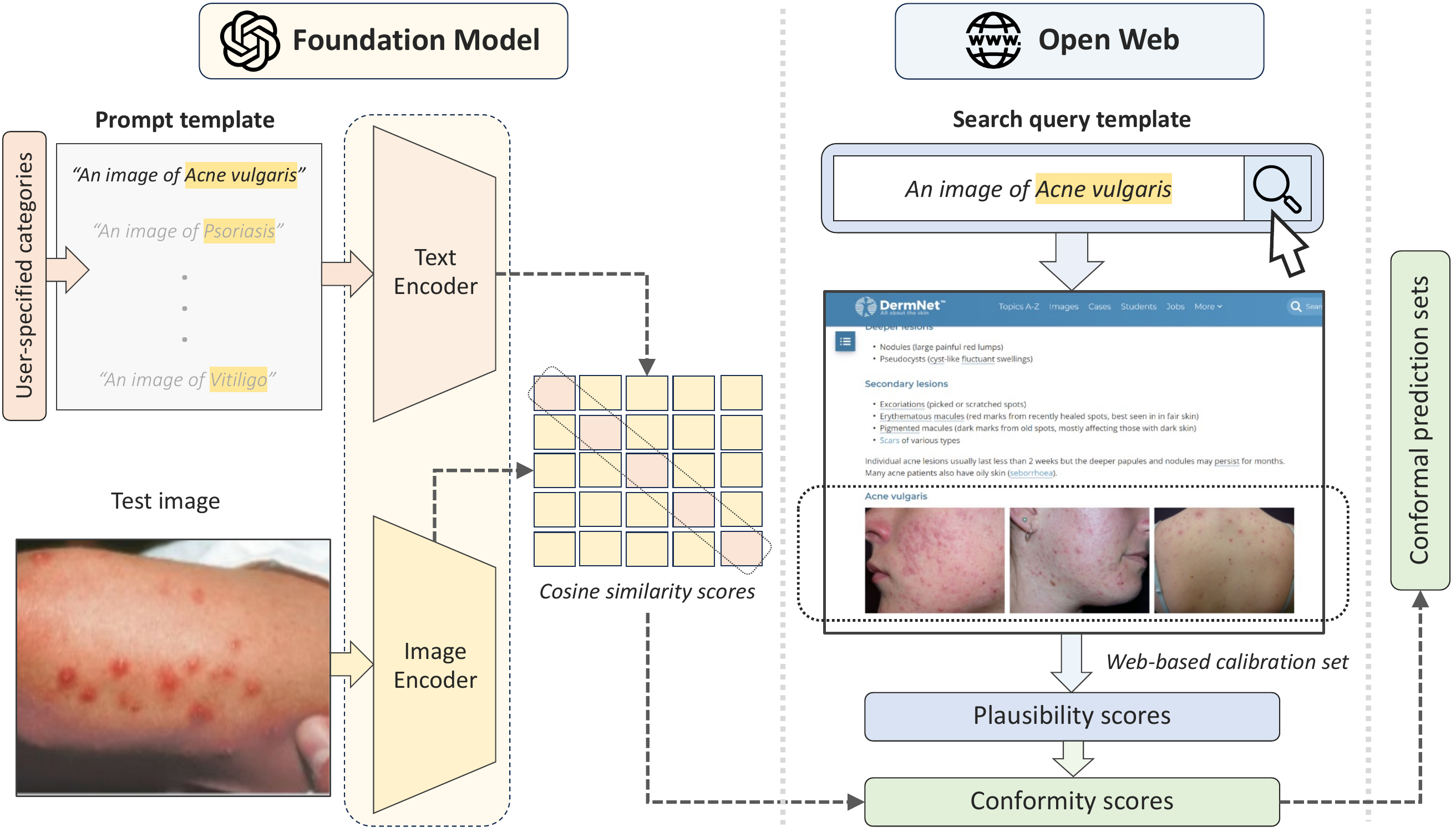}
  \caption{\footnotesize {\bf Illustration of our web-based conformal prediction procedure:} Given a test image and user-specified categories, we apply conformal prediction using ``on-the-fly'' calibration sets obtained~from~the~open~web.}
  \label{Fig1} 
\end{figure} 

\newpage
\section{Introduction}
Foundation models can be repurposed to tackle domain-specific use cases through zero-shot learning (ZSL) guided by task descriptions. In the ZSL paradigm, a pretrained foundation model~is~used~to~classify data according to previously unseen (user-specified) categories without the need for additional labeled examples for these categories. A common approach to ZSL is {\it contrastive~pretraining},~whereby encoders are trained to embed images and language into a shared, low-dimensional~latent~space~representing the semantic correspondence between visual and text data (\cite{rethmeier2023primer}). Models pretrained using this approach, such as CLIP, achieved remarkable~performance~in~a~broad range of computer vision benchmarks under zero-shot settings (\cite{Radford_Kim_Hallacy_Ramesh_Goh_Agarwal_Sastry_Askell_Mishkin_Clark,Cherti_Beaumont_Wightman_Wortsman_Ilharco_Gordon_Schuhmann_Schmidt_Jitsev_2022, Zhang_Xu_Usuyama_Bagga_Tinn_Preston_Rao_Wei_Valluri_Wong_et_al_2023}). {\bf The goal of this paper} is to develop methods for estimating uncertainty in zero-shot predictions based on pretrained foundation models. In line with~the~zero-shot~nature of these models, we seek versatile uncertainty estimation methods that can handle arbitrary user-specified categories and do not require access to labeled data specific to the task at hand.

Our proposed method for uncertainty estimation in foundation models is based on {\it conformal prediction} \citep{vovk2005algorithmic, Angelopoulos_gentle_intro}---a rigorous framework for predictive inference that can operate on top of any black-box model while providing distribution-free guarantees~on~the~coverage of its resulting prediction sets. The standard split conformal prediction (CP) procedure assumes access to a ``calibration set'' comprising labeled examples from the downstream task of interest. CP constructs prediction sets by evaluating the model errors on the held-out calibration set, and adjusting the set size so that it contains the true class in $1-\alpha$ of the calibration examples. If the calibration and testing data are drawn from the same distribution, CP is guaranteed to achieve a coverage probability of $1-\alpha$ in new test samples (\cite{vovk2005algorithmic}). However, in ZSL settings we do not have access to a predetermined set of categories, labeled calibration examples for these categories, or a known test distribution. Thus, to utilize CP for estimating uncertainty in foundation models within the ZSL setup, we need to develop new variants of the CP procedure that operate in a zero-shot fashion. 

{\bf Contributions.} In this paper, we develop a novel heuristic for zero-shot CP~that~operates~by~calibrating CLIP-based foundation models using data from the open web. The intuition behind our approach is that, in absence of predetermined classification categories and labeled calibration~data,~the~internet~can serve as a queryable source of universal calibration data that encompasses all possible user-specified categories and provides a reasonable approximation for downstream data distributions. Our web-based CP procedure operates at test time through the following steps: given a set of user-specified classes, we conduct zero-shot classification with CLIP-style models using a prompt template, e.g., {\it ``an image of a <category>''}, and use that prompt as a search query to source class-specific calibration data from the internet (Fig. \ref{Fig1}). Given the retrieved web-based calibration images and their associated contexts (i.e., HTML meta-data), we develop a procedure for generating ``plausibility scores'' that account for possible content or context errors in web search. These plausibility scores generated for each image and its associated context are heuristics for the probabilities of whether each particular user-given class is the ground-truth label for the image and its corresponding context. We then use these plausibility scores to conduct the Monte Carlo-based CP procedure proposed in (\cite{Stutz_2023}),~through~which~we obtain~prediction sets that capture the most likely classes for each test image.~We~evaluate~the~accuracy of our proposed method in multiple domain-specific image classification tasks, focusing on Biomedical datasets. We show that web-based CP empirically achieves target coverage while retaining efficiency comparable to that of an ``oracle'' CP procedure that uses data from target datasets for calibration.

\section{Related Work}
\vspace{-0.05in}
Given its model- and distribution-free properties, CP has been successfully applied to a wide range of applications, including calibration of Large Language Models (LLMs) (\citep{kumar2023conformal}) and traditional computer vision tasks \citep{andeol2023confident}. The closest work to ours is (\cite{kumar2022towards}), which also developed a zero-shot CP-based approach by framing the classification~task~as~an~outlier detection problem. This work proposes a different approach to zero-shot calibration---it~assumes~access to a held-out calibration set of image-caption pairs (drawn from the target~distribution),~and~proposes novel definitions of conformity scores based on the cosine similarity distance between images and captions. (\cite{fisch2021few}) proposes a few-shot CP approach that assumes availability of labeled calibration data for multiple related tasks, and exploits task similarity to derive prediction sets for new tasks for which limited data is available. Our CP procedure is based on the work in (\cite{Stutz_2023}), which proposes a calibration procedure for settings where the ground-truth is ambiguous. This procedure is where the concept of ``plausibility scores'' that reflect the level of ambiguity in labels originate from---our work proposes an incarnation of these scores for web-based calibration samples instead of based on MLE of expert opinion.

\section{Problem Formulation}
{\bf Setup and objectives.} We consider a classification problem where a context \( X \in \mathcal{X} \) is~assigned~an~associated label \( Y \in \mathcal{Y} \). Here, we define the label space \(\mathcal{Y}\) as a user-specified, finite~set~of~possible~categories relevant to an application of interest. Similarly, \(\mathcal{X}\) represents the set of all possible contexts that could be attributed to some label in \(\mathcal{Y}\). In vision-language foundation models,~the~context~space~\( \mathcal{X} \) could encompass all possible images that are compatible with some text label, or caption, within a specified set \( \mathcal{Y} \). Suppose we have access to a zero-shot multi-probability classifier~\mbox{\small \( \tau: \mathcal{X} \mapsto \Delta_\mathcal{Y} \)}, i.e., a foundation model, where \mbox{\small \( \Delta_\mathcal{Y} \subset \mathbf{R}^{\#|\mathcal{Y}|} \)} represents the probability simplex over \( \mathcal{Y} \). In other words, for a given context \( x \in \mathcal{X} \), the classifier output \( \tau(x) \) is a vector of probabilities, with each entry corresponding to the likelihood of each label \mbox{\small \( y \in \mathcal{Y} \)} being the ``true'' label for \( x \). An inherent challenge in ZSL is that it often requires probability predictions for out-of-distribution contexts and labels in $\mathcal{X} \times \mathcal{Y}$, which correspond to novel applications that did not appear in the training data of the foundation model $\tau$. Consequently, given a context $x \in \mathcal{X}$, the interpretation and usage of the multi-probability prediction  \mbox{\small $\tau(x) := \{ \tau(y \mid x): y \in \mathcal{Y}\} \in \Delta_\mathcal{Y}$} is unclear, as the reference distribution $P$ derived from the training data may not accurately or meaningfully reflect the novel prediction task of interest. To address this challenge, it is of interest to quantify uncertainty in zero-shot predictions through an application-specific prediction set \mbox{\small \( x \mapsto \widehat{C}(x) \subset \mathcal{Y} \)} that contains the true label with high probability. Specifically, for an application-specific ``ground-truth'' probability distribution \( \pi \) over \( \mathcal{X} \times \mathcal{Y} \), the prediction set \mbox{\small \( \widehat{C} \)} should satisfy the marginal coverage probability of \mbox{\small \(P_\pi(Y \in \widehat{C}(X)) = 1 - \alpha \)} for \mbox{\small \( \alpha \in (0,1) \)}, where \mbox{\small \(P_{\pi} \)} is the distribution of a test point \mbox{\small \( (X,Y) \sim P_\pi \)}. 

{\bf Conformal prediction.} When access to labeled data from the test distribution \mbox{\small \(P_{\pi} \)} is available,~CP~can be used to derive prediction sets \mbox{\small \( \widehat{C} \)} for which the coverage condition \mbox{\small \(P_\pi(Y \in \widehat{C}(X)) = 1 - \alpha \)}~is~guaranteed to hold \citep{vovk2005algorithmic}. The standard (split) CP procedure operates through the following steps. Given a labeled calibration set \mbox{\small \( \{(X_i, Y_i)\}_i\)}, we compute a {\it conformity score} \mbox{\small $V_i(\tau(X_i), Y_i)$} that measures the deviation of the prediction of the foundation model from the true label. A prediction set is then constructed by computing an empirical quantile of the conformity scores obtained from the labeled calibration set \mbox{\small $\{V_i\}_i$}. The definition of the conformity score depends on the application of interest. In the ZSL setup, we do not have access to the test distribution $P_\pi$ through a labeled calibration set \mbox{\small \( \{(X_i, Y_i)\}_i\)}, hence we cannot apply off-the-shelf CP. In the next section, we propose a zero-shot variant of CP in which the calibration set \mbox{\small \( \{(X_i, Y_i)\}_i\)} is sourced from the internet.

\section{Web-based Conformal Prediction with Foundation Models}
We propose an ``on-the-fly'' approach to CP in ZSL using data from~the~open web. As the open web contains images and contextual information from diverse~open sources (e.g., social media, photo galleries, scientific reporting), it serves as a rich and easily-accessible universal~calibration~dataset~that~is applicable to most domains. Our CP procedure, dubbed WebCP, involves the steps below. 

{\bf Step 1: Calibration data mining.} Given a user-specified set of categories \mbox{\small \( \mathcal{Y}\)}, the first step in~WebCP~is to acquire a set of image and web meta-data corresponding to these categories, i.e.,
\begin{align}
\mathcal{C}_\text{web} = \{(\widetilde{X}^{y_i}_i, M^{y_i}_i): i \in [1,\ldots, \textstyle \sum_{k=1}^{\#({\mathcal{Y}})} n_{y_k}]\}.
\label{NEQ1}
\end{align}
Here, for each sample $i$ in $\mathcal{C}_\text{web}$,  \mbox{\small $\widetilde{X}^{y_i}_i$} is an image corresponding to the class $y_i$ downloaded from a web source (e.g., Google search) and \mbox{\small $M^{y_i}_i$} is its accompanied web meta-data (e.g., textual content of the web page source HTML). We obtain the images \mbox{\small $\{\widetilde{X}^{y_i}_i\}_i$} by querying a web source 
 for each class \mbox{\small $y_k \in \mathcal{Y}$} with a template query that depends on the particular class $y_k$, collecting~the~first \mbox{\small $n_{y_k}$} image search results for each \mbox{\small $y_k \in \mathcal{Y}$}, and aggregating them together to construct the calibration set in (Equation \ref{NEQ1}). Note our use of potentially different template prompts; we use a template prompt (e.g., {\it ``An image of $y_k$''}) to evaluate the predictions of a CLIP-based foundation model, and a template search query (which could be the same as the template prompt) to source~calibration~data from the internet. For instance, if $y_k$ is a skin condition such as {\it Acne Vulgaris}~in~a~dermatology~application, then the prompt and search query can be {\it ``An image of Acne Vulgaris''}~(Fig.~\ref{Fig1}). The meta-data \mbox{\small $M^{y_i}_i$} associated with image \mbox{\small $\widetilde{X}^{y_i}_i$} comprises textual context associated with the web page from which the image is obtained. In our setup, we extract textual meta-data retrieving the content of the {\color{purple}{\texttt{<alt>}}}~tag~and the text around the {\color{purple}{\texttt{<img>}}} header within the HTML source code of the web page hosting the image \mbox{\small $\widetilde{X}^{y_i}_i$}. Full details of our web scarping procedure are provided in the Supplementary Appendix (\ref{app:data_mining}). 

{\bf Step 2: Plausibility Generation.} For each mined image in (\ref{NEQ1}), we estimate a collection of ``plausibility''~scores~defined as \mbox{\small $\boldsymbol{\lambda}_i = \{\lambda^y_i: y \in \mathcal{Y}\}$}. Each plausibility score \mbox{\small $\lambda^y_i$} in the vector $\boldsymbol{\lambda}_i$ is estimated based on the meta-data~\mbox{\small $M^{y_i}_i$} and the contents of the image \mbox{\small $\widetilde{X}_i^{y_i}$},~and~reflects the likelihood that the retrieved image \mbox{\small $\widetilde{X}^{y_i}_i$} belongs to the each of the particular user-specified classes \mbox{\small $y \in \mathcal{Y}$}.~Our~definition of plausibility scores captures two forms of alignment between the search query and retrieved results: {\bf context alignment} and {\bf content alignment}. The context alignment between a sampled image $\widetilde{X}^{y_i}_i$ and any class $y \in \mathcal{Y}$ reflects the relevance of the source web page and its meta-data for that image \mbox{\small $M^{y_i}_i$} to the search query for $y$ ({\it ``An image of $y$''}). We quantify context alignment through a heuristic that assesses the relevance of the textual content of the source web page to the provided search query. On the other hand, content alignment reflects the relevance of the image embedded in the web page to the query corresponding to the class $y$. Content misalignment occurs when the image doesn't generically fit the search query, or if the image is topically~accurate~but~is displayed in an undesired form (e.g., a cartoon illustration~or~a~diagram~of~the~queried~class~$y$).~We quantify content alignment via Content-Based Image Retrieval (CBIR) methods (\cite{muller2001performance}).

Each of the plausibility scores contained in the vector \mbox{\small $ \boldsymbol{\lambda}_i = \{ \lambda^y_i: y \in \mathcal{Y}\}$} generated for each image \mbox{\small $\widetilde{X}^{y_i}_i$} are computed by combining two~components:~context~alignment scores and content alignment scores. The context alignment score measures context alignment through an algorithm \mbox{\small \texttt{context}($M^{y_i}_i$, $y$)}; this algorithm evaluates the relevance of the web page meta-data \mbox{\small $M^{y_i}_i$} to a given class \mbox{\small $y$} by feeding it into a text encoder (Sentence-BERT,  \cite{Reimers_Gurevych_2019}) to obtain sentence-level embeddings of \mbox{\small $M^{y_i}_i$} which are then compared to the search query embedding (label name) of $y$ via cosine similarity. For each image $\widetilde{X}_i^{y_i}$, the resulting scores for the image across all classes $y \in \mathcal{Y}$ are normalized using a temperature-tuned softmax to obtain context alignment scores $\{c_i^y: y \in \mathcal{Y}\}$ for that image $\widetilde{X}_i^{y_i}$. 

 To evaluate content alignment scores, we run each image through an algorithm \mbox{\small \texttt{content}($X^{y_i}_i$, $y$)} to evaluate the probability that the retrieved image is not depicting the queried category. This algorithm includes two stages: (i) a filtering stage for detecting when an image is topically relevant but not of the correct form such as images of diagrams or charts, and (ii) a scoring stage to check if the image contains a general visual representation of a class \mbox{\small $y \in \mathcal{Y}$}. 
 
 For the filtering stage, we utilized CLIP to generate similarity scores between the image and a series of invalid form prompts (e.g. \textit{"an image with a lot of text"}, \textit{"an image of a graph"}) alongside a "negative" label (\textit{"an image"}) used to filter out "invalid" images. The idea is that if the image is similar to one of the invalid form prompts (which were formulated through experimentation) then it is of an invalid form. As such, we take the softmax of the resulting scores and use the negative label score ${s_\text{neg}}_i$ as the probability that the image is of a valid form. Note that ${s_\text{neg}}_i$ is generated for each image without considering the classes in $\mathcal{Y}$; this is because the negative label score reflects the likelihood of an image being of an "invalid" form which does not depend on any of the classes $y \in \mathcal{Y}$.

For the scoring stage, we wish to analyze image-based content alignment and detect content misalignment errors for the image across each class $y \in \mathcal{Y}$; we do this by applying CLIP to a relaxed version of the original classification task (i.e. determining if $\widetilde{X}_i^{y_i}$ conforms to a class $y \in \mathcal{Y}$) to detect if the image doesn't fit the high-level visual archetype of the label given for $y$. To do this we first have to generate a set of simplified variants of the class labels \mbox{\small $\{y: y \in \mathcal{Y}\}$} that act as generic visual representations of a target label. Ideally these generalized variants are a) high-level enough to ensure that the zero-shot classifier can distinguish them with a high probability and b) representative of the obvious visual features in the image. To obtain this simplified variant \mbox{\small $y_\text{pseudo}$} for a corresponding $y \in \mathcal{Y}$, we use a named entity recognition model to identify all entities in a tag, reference these entities to an ontology (e.g. Dbpedia, \cite{Auer_Bizer_Kobilarov_Lehmann_Cyganiak_Ives_2007}; MeSH, \cite{Rogers_1963}; etc.), and then substitute the entity with a term from higher in the hierarchy. For instance, if \mbox{\small $y=$}\textit{``colorectal adenocarcinoma epithelium''} then a simplified label could be \textit{``microscope image''}. Once the generalized labels are derived, we generate softmax scores for each simplified label in comparison to the negative label \textit{"an image"}, generating a set of content alignment scores $\{h_i^{y}: y \in \mathcal{Y}\}$ for each image $\widetilde{X}_{i}^{y_i} \in \widetilde{\mathcal{C}}_\text{web}$. 

 To integrate context/content alignments in generating plausibility scores $\boldsymbol{\lambda}_i = \{\lambda_i^y: y \in \mathcal{Y}\}$ corresponding to a sample $(\widetilde{X}^{y_i}_i, M^{y_i}_i)$ that represents the likelihood of that sample belonging to classes $y \in \mathcal{Y}$, we take a pairwise product of the context alignment scores \mbox{\small $\{c_i^y: y \in \mathcal{Y}\}$} and the content alignment scores \mbox{\small $\{h_i^y: y \in \mathcal{Y}\}$}, and scale each of the computed products by ${s_\text{neg}}_i$; thus, we have $\lambda_i^y = c_i^y \, h_i^y \, {s_\text{neg}}_i$ for each $y \in \mathcal{Y}$. Then, to reflect the possibility of a sampled image being irrelevant to any of the classes $y \in \mathcal{Y}$, we subtract the summation of the generated class probabilities $\lambda_i^y$ across all $y \in \mathcal{Y}$ from $1.0$ to derive a so-called "junk probability". To summarize, we have:
 \begin{alignat}{3}
     \lambda_i^y &= h_i^y \, c_i^y \, s_\text{neg},\,\, {\lambda_\text{junk}}_i &= 1-\sum_{i=1}^{n}\lambda_i^y.
 \end{alignat} 
 The resulting classifier scores are then used as estimates of content plausibility. We utilize the plausibility scores to construct the final calibration set as follows: 
\begin{align}
\widetilde{\mathcal{C}}_\text{web} = \{(\widetilde{X}^{y_i}_i, \boldsymbol{\lambda}_i, {\lambda_{\text{junk}}}_i): i \in [1,\ldots,n_y]\}.
\label{NEQ2}
\end{align}
An overview of this discussion and the \mbox{\small \texttt{context}} and \mbox{\small \texttt{content}} algorithms we choose to use are provided in relevant figures in the Appendix (\ref{app:overview}).

\vspace{-.075in}
\begin{algorithm}[!h]
{\footnotesize
\caption{Web-based Conformal Prediction (WebCP)}
\label{alg:mccp_modified}
\vspace{.025in}
\textbf{Input:} User-specified classes \mbox{\footnotesize $\mathcal{Y}$}; coverage \mbox{\footnotesize $1-\alpha$}; Monte Carlo samples \mbox{\footnotesize $M$}; foundation model  \mbox{\footnotesize $\tau$}; test image \mbox{\footnotesize $X$}
\vspace{-.025in}
\begin{enumerate}
\item {\bf Calibration data mining:} Download images and meta-data pairs \mbox{\footnotesize $\{(\widetilde{X}^{y_i}_i, M^{y_i}_i)\}_{i}$} from an online source using a search query template filled with the categories in \mbox{\footnotesize $\mathcal{Y}$}. 
\item {\bf Plausibility generation:} Estimate context and content alignment of the scraped images~\mbox{\footnotesize $\{\widetilde{X}^{y_i}_i\}_{i}$}~using the \mbox{\footnotesize \texttt{context}($M^{y_i}_i$, $y$)} and \mbox{\small \texttt{content}($\widetilde{X}^{y_i}_i$, $y$)} algorithms to estimate the plausibility scores \mbox{\footnotesize $\{(\boldsymbol{\lambda}_i, {\textstyle \lambda_{\text{junk}}}_i)\}_i$}.
\item {\bf Monte Carlo CP:} Perform sampling $M$ times, where on each iteration $m \in [1,\ldots, M]$:
\begin{itemize}
\item Iterate through each calibration example \mbox{\footnotesize $(\widetilde{X}^{y_i}_i, \boldsymbol{\lambda}_i, {\lambda_{\text{junk}}}_i)$}. Choose to reject the example with probability ${\lambda_{\text{junk}}}_i$, or keep it with probability $1-{\lambda_{\text{junk}}}_i$. If kept, randomly sample a label \mbox{\footnotesize $\widetilde{y}_i$} in \mbox{\footnotesize $\mathcal{Y}$} from the distribution \mbox{\footnotesize $\widetilde{y}_i \sim$ Categorical($\lambda^y_i: y \in \mathcal{Y}$)}, and add the example $(\widetilde{X}^{y_i}_i, \widetilde{y}^m_i)$ to the random calibration set for the current iteration. Our final random calibration set for this iteration will be \mbox{\footnotesize $\widetilde{\mathcal{C}}_m = \{(\widetilde{X}^{y_i}_i, \widetilde{y}^m_i)\}_i$}.
\item Using the aggregate calibration dataset \mbox{\footnotesize $\widetilde{\mathcal{C}} = \{\widetilde{\mathcal{C}}_1, \ldots, \widetilde{\mathcal{C}}_M\}$} find the minimum threshold $\gamma$ s.t.
\end{itemize}
$$\frac{1}{M} \sum_{m=1}^M \left[\frac{\sum_{m^\prime=1}^{|\widetilde{\mathcal{C}}_{m^\prime}|} \mathbbm{1}\{V(\tau(\widetilde{X}^y_{m^\prime}), \widetilde{y}^m_{m^\prime}) \leq \gamma\} + 1}{|\widetilde{\mathcal{C}}_{m}| + 1} \right] > 1-\alpha$$.
\item {\bf Prediction set construction:} For a new test image \mbox{\footnotesize $X$}, \mbox{\footnotesize $\mathcal{P}(X) = \{y \in \mathcal{Y}: V(\tau(X), y) \leq \gamma\}.$}
\end{enumerate}
\textbf{Output:} A prediction set \mbox{\footnotesize $\mathcal{P}(X) \subseteq \mathcal{Y}$}.}
\vspace{.025in}
\end{algorithm}
\vspace{-.05in}
{\bf Step 3: Conformal Prediction with Ambiguous Ground-truths.} The calibration set in (\ref{NEQ2}) comprises a set of images for each class along with probabilistic (ambiguous) labels for memberships to each class in \mbox{\small $\mathcal{Y}$}. To construct CP-based prediction sets for new test images, we apply the Monte Carlo CP procedure proposed in (\cite{Stutz_2023}), which accounts for ambiguity in ground-truth labels, to the calibration set \mbox{\small $\widetilde{\mathcal{C}}_\text{web}$}. For our conformity score, we use the softmax transformed cosine similarity score between the CLIP image embedding and the CLIP label/caption embedding. A pseudo-code for the overall WebCP procedure~is~provided~in~Algorithm~1.

{\bf The internet as a universal CP calibration set.} One of the key advantages of CP is that it provides provable guarantees on the coverage of the prediction sets \mbox{\footnotesize $\mathcal{P}$} in test data. However,~WebCP~is~a~heuristic that sources on-the-fly calibration data from a large knowledge base (i.e., the internet) in order to calibrate zero-shot predictions, without access to the test distribution of the downstream task. This means that the exchangeability assumption, a necessary condition for the CP coverage guarantees (\cite{vovk2005algorithmic}), no longer holds. Consequently, the central question of interest is whether WebCP can attain the desired target coverage levels---a question that now hinges on empirical validation. In other words, we would like to test if web-scraped calibration data \mbox{\small $\widetilde{\mathcal{C}}_\text{web} \sim P_\text{web}$} is a good approximation of oracle calibration sets drawn from the test distribution \mbox{\small $\widetilde{\mathcal{C}}^* \sim P_{\pi}$} across many domain-specific datasets. In the next Section, we test this hypothesis within the context of biomedical applications. 

\begin{table}[h]
            \scriptsize
            \caption{{\scriptsize $\Delta_{cov}$ = Difference between Target (1-$\alpha$) and Achieved Coverage}}
            \label{results-table}
            \centering
            \begin{tabular}{@{}lllllcllllc@{}}
            \toprule
                                  & \multicolumn{5}{c}{\textbf{MedMNIST Microscopy Subset}}                                                                                 & \multicolumn{5}{c}{\textbf{FitzPatrick-17k}}                                                                                                       \\ \midrule
                                  & \multicolumn{2}{c}{Calibration}       & \multicolumn{2}{c}{Test}                    & \multicolumn{1}{l}{}                        & \multicolumn{2}{c}{Calibration}             & \multicolumn{2}{c}{Test}                    & \multicolumn{1}{l}{}                                   \\
            \multicolumn{1}{c}{$\alpha$} & Coverage                       & Efficiency & Coverage                       & Efficiency & \multicolumn{1}{c}{$\Delta_{cov}$} & Coverage                       & Efficiency & Coverage                       & Efficiency & \multicolumn{1}{c}{$\Delta_{cov}$}            \\ \cmidrule(lr){2-5} \cmidrule(lr){7-10}
            \multicolumn{11}{l}{\textbf{WebCP}}                                                                                                                                                                                                                                                     \\ \midrule
            \textbf{0.10}             & \cellcolor[HTML]{EFEFEF}0.9643 & 13.41      & \cellcolor[HTML]{EFEFEF}0.9823 & 18.28      & \cellcolor[HTML]{9AFF99}8.23\%              & \cellcolor[HTML]{EFEFEF}0.9253 & 75.51      & \cellcolor[HTML]{EFEFEF}0.9023 & 78.67      & \cellcolor[HTML]{9AFF99}{\color[HTML]{000000} +0.20\%} \\
            \textbf{0.20}             & \cellcolor[HTML]{EFEFEF}0.8791 & 10.18      & \cellcolor[HTML]{EFEFEF}0.8473 & 14.38      & \cellcolor[HTML]{9AFF99}4.73\%              & \cellcolor[HTML]{EFEFEF}0.8381 & 51.40      & \cellcolor[HTML]{EFEFEF}0.8007 & 54.96      & \cellcolor[HTML]{9AFF99}{\color[HTML]{000000} +0.04\%} \\
            \textbf{0.30}             & \cellcolor[HTML]{EFEFEF}0.7800 & 7.944      & \cellcolor[HTML]{EFEFEF}0.7236 & 11.26      & \cellcolor[HTML]{9AFF99}2.37\%              & \cellcolor[HTML]{EFEFEF}0.7497 & 37.15      & \cellcolor[HTML]{EFEFEF}0.7064 & 40.33      & \cellcolor[HTML]{9AFF99}{\color[HTML]{000000} +0.64\%} \\
            \textbf{0.40}             & \cellcolor[HTML]{EFEFEF}0.6939 & 6.53       & \cellcolor[HTML]{EFEFEF}0.6423 & 9.29       & \cellcolor[HTML]{9AFF99}4.23\%              & \cellcolor[HTML]{EFEFEF}0.6510 & 25.33      & \cellcolor[HTML]{EFEFEF}0.6028 & 27.90      & \cellcolor[HTML]{9AFF99}{\color[HTML]{000000} +0.28\%} \\
            \textbf{0.50}             & \cellcolor[HTML]{EFEFEF}0.5819 & 4.94       & \cellcolor[HTML]{EFEFEF}0.5163 & 6.81       & \cellcolor[HTML]{9AFF99}1.63\%              & \cellcolor[HTML]{EFEFEF}0.5445 & 16.46      & \cellcolor[HTML]{EFEFEF}0.4920 & 18.29      & \cellcolor[HTML]{FFCCC9}-0.79\%                        \\ \cmidrule(lr){2-5} \cmidrule(lr){7-10}
            \multicolumn{11}{l}{\textbf{Standard CP with web-based calibration data}}                                                                                                                                                                                                                                                      \\ \midrule
            \textbf{0.10}             & \cellcolor[HTML]{EFEFEF}0.9009 & 10.71      & \cellcolor[HTML]{EFEFEF}0.8713 & 15.05      & \cellcolor[HTML]{FFCCC9}-2.87\%             & \cellcolor[HTML]{EFEFEF}0.9002 & 67.23      & \cellcolor[HTML]{EFEFEF}0.8738 & 70.64      & \cellcolor[HTML]{FFCCC9}-2.61\%                        \\
            \textbf{0.20}             & \cellcolor[HTML]{EFEFEF}0.8003 & 8.39       & \cellcolor[HTML]{EFEFEF}0.7503 & 11.90      & \cellcolor[HTML]{FFCCC9}-4.97\%             & \cellcolor[HTML]{EFEFEF}0.8001 & 44.59      & \cellcolor[HTML]{EFEFEF}0.7585 & 48.05      & \cellcolor[HTML]{FFCCC9}-4.14\%                        \\
            \textbf{0.30}             & \cellcolor[HTML]{EFEFEF}0.7004 & 6.62       & \cellcolor[HTML]{EFEFEF}0.6470 & 9.43       & \cellcolor[HTML]{FFCCC9}-5.29\%             & \cellcolor[HTML]{EFEFEF}0.7001 & 30.82      & \cellcolor[HTML]{EFEFEF}0.6565 & 33.72      & \cellcolor[HTML]{FFCCC9}-4.35\%                        \\
            \textbf{0.40}             & \cellcolor[HTML]{EFEFEF}0.6006 & 5.15       & \cellcolor[HTML]{EFEFEF}0.5316 & 7.10       & \cellcolor[HTML]{FFCCC9}-6.84\%             & \cellcolor[HTML]{EFEFEF}0.6000 & 20.73      & \cellcolor[HTML]{EFEFEF}0.5492 & 22.97      & \cellcolor[HTML]{FFCCC9}-5.08\%                        \\
            \textbf{0.50}             & \cellcolor[HTML]{EFEFEF}0.5000 & 3.85       & \cellcolor[HTML]{EFEFEF}0.4317 & 5.20       & \cellcolor[HTML]{FFCCC9}-6.83\%             & \cellcolor[HTML]{EFEFEF}0.5000 & 13.44      & \cellcolor[HTML]{EFEFEF}0.4433 & 14.90      & \cellcolor[HTML]{FFCCC9}-5.66\%                        \\ \cmidrule(lr){2-5} \cmidrule(lr){7-10}
            \multicolumn{11}{l}{\textbf{Oracle CP with calibration on target data}}                                                                                                                                                                                                                                            \\ \midrule
            \textbf{0.10}             & \cellcolor[HTML]{EFEFEF}0.9180 & 11.28      & \cellcolor[HTML]{EFEFEF}0.9072 & 15.94      & \cellcolor[HTML]{9AFF99}0.72\%              & \cellcolor[HTML]{EFEFEF}0.9199 & 73.49      & \cellcolor[HTML]{EFEFEF}0.8933 & 76.62      & \cellcolor[HTML]{FFCCC9}-0.66\%                        \\
            \textbf{0.20}             & \cellcolor[HTML]{EFEFEF}0.8498 & 9.36       & \cellcolor[HTML]{EFEFEF}0.8031 & 13.28      & \cellcolor[HTML]{9AFF99}0.32\%              & \cellcolor[HTML]{EFEFEF}0.8380 & 51.53      & \cellcolor[HTML]{EFEFEF}0.8024 & 55.01      & \cellcolor[HTML]{9AFF99}0.24\%                         \\
            \textbf{0.30}             & \cellcolor[HTML]{EFEFEF}0.7613 & 7.67       & \cellcolor[HTML]{EFEFEF}0.7053 & 10.93      & \cellcolor[HTML]{9AFF99}0.53\%              & \cellcolor[HTML]{EFEFEF}0.7530 & 37.59      & \cellcolor[HTML]{EFEFEF}0.7143 & 40.72      & \cellcolor[HTML]{9AFF99}1.43\%                         \\
            \textbf{0.40}             & \cellcolor[HTML]{EFEFEF}0.6582 & 5.98       & \cellcolor[HTML]{EFEFEF}0.6001 & 8.52       & \cellcolor[HTML]{9AFF99}0.11\%              & \cellcolor[HTML]{EFEFEF}0.6520 & 25.43      & \cellcolor[HTML]{EFEFEF}0.6057 & 27.99      & \cellcolor[HTML]{9AFF99}0.57\%                         \\
            \textbf{0.50}             & \cellcolor[HTML]{EFEFEF}0.5698 & 4.74       & \cellcolor[HTML]{EFEFEF}0.5045 & 6.60       & \cellcolor[HTML]{9AFF99}0.45\%              & \cellcolor[HTML]{EFEFEF}0.5557 & 17.24      & \cellcolor[HTML]{EFEFEF}0.5055 & 19.13      & \cellcolor[HTML]{9AFF99}0.55\%                         \\ \bottomrule
            \end{tabular}
        \end{table}

\section{Experiments}
{\bf Experimental setup.} We evaluate the WebCP procedure biomedical datasets~with~variants~of~CLIP as the underlying multimodal foundation model. Specifically, we evaluate the effectiveness of WebCP at generating efficient and prediction sets for the black-box BioMedCLIP model {\color{purple}{\texttt{microsoft/BiomedCLIP-PubMedBERT\_256-vit\_base\_patch16\_224}}} (\cite{microsoft/BiomedCLIP-PubMedBERT_256-vit_base_patch16_224_Hugging_Face}), which is a foundation model that uses PubMedBERT as the text encoder and~is~finetuned~on biomedical tasks by pre-training on PMC-15M (\cite{Zhang_Xu_Usuyama_Bagga_Tinn_Preston_Rao_Wei_Valluri_Wong_et_al_2023}). To estimate context~alignment,~we~use the Sentence-BERT model {\color{purple}{\texttt{sentence-transformers/msmarco-bert-base-dot-v5}}} which was finetuned on MS-MARCO (\cite{sentence-transformers/msmarco-bert-base-dot-v5_Hugging_Face, Bajaj_Campos_Craswell_Deng_Gao_Liu_Majumder_McNamara_Mitra_Nguyen_et_al_2018})). For content alignment estimation, we use a variant of CLIP finetuned on the LAION dataset (\cite{laion/CLIP-convnext_large_d.laion2B-s26B-b102K-augreg_hugging_face}). The reason why we use different variants of CLIP for content alignment and classification is two-fold. Firstly, the simplified labels should be very generic (i.e. "dog", "island", etc.) and as such a base CLIP model, which has been shown to perform reasonably well in general image classification benchmarks such as ImageNet, should provide satisfactory estimates for this task (\cite{Radford_Kim_Hallacy_Ramesh_Goh_Agarwal_Sastry_Askell_Mishkin_Clark}). Secondly, using different variants of CLIP helps to decorrelate errors between the classification and plausibility tasks, thereby decreasing the probability of overconfidence due to matched errors in the two tasks.

To acquire the calibration data, we use a Selenium-based web crawling agent and Google Custom Image Search Engine to query the dataset-specific image classes, and cache the images and corresponding captions from the top 50 search results (after filtering out those missing textual contexts). We compare our proposed WebCP method with two baseline procedures. The first is a {\it Standard CP} procedure applied to the mined calibration data, without accounting for ambiguity in image classes through the generated plausibility scores. The second is an {\it Oracle CP} procedure, which applies the standard CP calibration step to an equal number of samples drawn from the target dataset on which coverage is evaluated. We evaluate all baselines in terms of their achieved coverage on an unseen test set as well as their efficiency, i.e., the average size of the prediction set \mbox{\footnotesize $|\mathcal{P}(X)|$}.

{\bf Datasets.} We consider two biomedical datasets: (1) \texttt{Fitzpatrick17k}, a dataset containing annotated medical images of 114 classes of skin conditions \cite{groh2021evaluating}, and (2) the \texttt{PathMNIST}, \texttt{BloodMNIST}, and \texttt{TissueMNIST} subsets from the \texttt{MedMNIST} dataset~(\cite{yang2023medmnist}),~which~contains an aggregation of low-resolution images under 25 classes of image types derived from studies in colon pathology, blood cell microscope, and kidney cortex microscope imaging.

{\bf Results.} In almost all experiments, our WebCP procedure results in prediction sets that consistently achieves the targeted $1-\alpha$ coverage on the test datasets, and displays~satisfactory~calibration~performance across varying levels of $\alpha$. On the contrary, the Standard CP procedure applies to the scraped calibration data significantly under-covers in the test dataset across all values of $\alpha$. This shows the utility of our generated plausibility scores, which accounts for potential content and context errors in the retrieved web data. It also shows that, for the datasets under consideration, web-scraped data can provide useful calibration sets that can help calibrate the zero-shot predictions of foundation models. Compared to the Oracle CP, WebCP incurs a slight loss of efficiency, which is expected due to the conservative nature of calibration under ambiguous ground truth. However, the efficiency of WebCP remains comparable to the oracle efficiency for all values of $\alpha$ across the two datasets.

\section{Conclusion}
In this paper, we developed a method for estimating uncertainty in~the~zero-shot~predictions~of~pretrained foundation models. Our proposed heuristic estimates uncertainty in image classification using conformal prediction applied to web-scraped data in a zero-shot fashion. Our procedure,~dubbed~WebCP, comprises three steps: (1) mining calibration data based on user-specified classification categories, (2) estimating plausibility scores to quantify the alignment of the mined data with the user queries, and (3) applying a Monte Carlo-based approach to conformal prediction using the estimated plausibility scores in order to calibrate the predictions of foundation models. Our preliminary results show that WebCP could be a promising approach to zero-shot calibration---in biomedical datasets, it achieves the user-specified target coverage while retaining competitive efficiency on test data.

\bibliographystyle{iclr2023_conference}
\bibliography{neurips2023_submission}

\begin{thebibliography}{23}
\providecommand{\natexlab}[1]{#1}
\providecommand{\url}[1]{\texttt{#1}}
\expandafter\ifx\csname urlstyle\endcsname\relax
  \providecommand{\doi}[1]{doi: #1}\else
  \providecommand{\doi}{doi: \begingroup \urlstyle{rm}\Url}\fi

\bibitem[And{\'e}ol et~al.(2023)And{\'e}ol, Fel, De~Grancey, and
  Mossina]{andeol2023confident}
L{\'e}o And{\'e}ol, Thomas Fel, Florence De~Grancey, and Luca Mossina.
\newblock Confident object detection via conformal prediction and conformal
  risk control: an application to railway signaling.
\newblock \emph{arXiv preprint arXiv:2304.06052}, 2023.

\bibitem[Angelopoulos \& Bates(2022)Angelopoulos and
  Bates]{Angelopoulos_gentle_intro}
Anastasios~N. Angelopoulos and Stephen Bates.
\newblock A gentle introduction to conformal prediction and distribution-free
  uncertainty quantification.
\newblock \penalty0 (arXiv:2107.07511), December 2022.
\newblock \doi{10.48550/arXiv.2107.07511}.
\newblock URL \url{http://arxiv.org/abs/2107.07511}.
\newblock arXiv:2107.07511 [cs, math, stat].

\bibitem[Auer et~al.(2007)Auer, Bizer, Kobilarov, Lehmann, Cyganiak, and
  Ives]{Auer_Bizer_Kobilarov_Lehmann_Cyganiak_Ives_2007}
Sören Auer, Christian Bizer, Georgi Kobilarov, Jens Lehmann, Richard Cyganiak,
  and Zachary Ives.
\newblock Dbpedia: A nucleus for a web of open data.
\newblock In Karl Aberer, Key-Sun Choi, Natasha Noy, Dean Allemang, Kyung-Il
  Lee, Lyndon Nixon, Jennifer Golbeck, Peter Mika, Diana Maynard, Riichiro
  Mizoguchi, Guus Schreiber, and Philippe Cudré-Mauroux (eds.), \emph{The
  Semantic Web}, Lecture Notes in Computer Science, pp.\  722–735, Berlin,
  Heidelberg, 2007. Springer.
\newblock ISBN 978-3-540-76298-0.
\newblock \doi{10.1007/978-3-540-76298-0_52}.

\bibitem[Bajaj et~al.(2018)Bajaj, Campos, Craswell, Deng, Gao, Liu, Majumder,
  McNamara, Mitra, Nguyen, Rosenberg, Song, Stoica, Tiwary, and
  Wang]{Bajaj_Campos_Craswell_Deng_Gao_Liu_Majumder_McNamara_Mitra_Nguyen_et_al_2018}
Payal Bajaj, Daniel Campos, Nick Craswell, Li~Deng, Jianfeng Gao, Xiaodong Liu,
  Rangan Majumder, Andrew McNamara, Bhaskar Mitra, Tri Nguyen, Mir Rosenberg,
  Xia Song, Alina Stoica, Saurabh Tiwary, and Tong Wang.
\newblock Ms marco: A human generated machine reading comprehension dataset.
\newblock \penalty0 (arXiv:1611.09268), Oct 2018.
\newblock URL \url{http://arxiv.org/abs/1611.09268}.
\newblock arXiv:1611.09268 [cs].

\bibitem[Bird et~al.(2009)Bird, Klein, and Loper]{Bird_Klein_Loper_2009}
Steven Bird, Ewan Klein, and Edward Loper.
\newblock \emph{Natural Language Processing with Python: Analyzing Text with
  the Natural Language Toolkit}.
\newblock O’Reilly Media, Beijing.; Cambridge Mass., 1st edition edition, Aug
  2009.
\newblock ISBN 978-0-596-51649-9.

\bibitem[Cherti et~al.(2022)Cherti, Beaumont, Wightman, Wortsman, Ilharco,
  Gordon, Schuhmann, Schmidt, and
  Jitsev]{Cherti_Beaumont_Wightman_Wortsman_Ilharco_Gordon_Schuhmann_Schmidt_Jitsev_2022}
Mehdi Cherti, Romain Beaumont, Ross Wightman, Mitchell Wortsman, Gabriel
  Ilharco, Cade Gordon, Christoph Schuhmann, Ludwig Schmidt, and Jenia Jitsev.
\newblock Reproducible scaling laws for contrastive language-image learning.
\newblock \penalty0 (arXiv:2212.07143), Dec 2022.
\newblock URL \url{http://arxiv.org/abs/2212.07143}.
\newblock arXiv:2212.07143 [cs].

\bibitem[Fisch et~al.(2021)Fisch, Schuster, Jaakkola, and
  Barzilay]{fisch2021few}
Adam Fisch, Tal Schuster, Tommi Jaakkola, and Regina Barzilay.
\newblock Few-shot conformal prediction with auxiliary tasks.
\newblock In \emph{International Conference on Machine Learning}, pp.\
  3329--3339. PMLR, 2021.

\bibitem[Groh et~al.(2021)Groh, Harris, Soenksen, Lau, Han, Kim, Koochek, and
  Badri]{groh2021evaluating}
Matthew Groh, Caleb Harris, Luis Soenksen, Felix Lau, Rachel Han, Aerin Kim,
  Arash Koochek, and Omar Badri.
\newblock Evaluating deep neural networks trained on clinical images in
  dermatology with the fitzpatrick 17k dataset.
\newblock In \emph{Proceedings of the IEEE/CVF Conference on Computer Vision
  and Pattern Recognition}, pp.\  1820--1828, 2021.

\bibitem[Kumar et~al.(2022)Kumar, Palepu, Tuwani, and Beam]{kumar2022towards}
Bhawesh Kumar, Anil Palepu, Rudraksh Tuwani, and Andrew Beam.
\newblock Towards reliable zero shot classification in self-supervised models
  with conformal prediction.
\newblock \penalty0 (arXiv:2210.15805), Oct 2022.
\newblock \doi{10.48550/arXiv.2210.15805}.
\newblock URL \url{http://arxiv.org/abs/2210.15805}.
\newblock arXiv:2210.15805 [cs].

\bibitem[Kumar et~al.(2023)Kumar, Lu, Gupta, Palepu, Bellamy, Raskar, and
  Beam]{kumar2023conformal}
Bhawesh Kumar, Charlie Lu, Gauri Gupta, Anil Palepu, David Bellamy, Ramesh
  Raskar, and Andrew Beam.
\newblock Conformal prediction with large language models for multi-choice
  question answering.
\newblock \emph{arXiv preprint arXiv:2305.18404}, 2023.

\bibitem[M{\"u}ller et~al.(2001)M{\"u}ller, M{\"u}ller, Squire,
  Marchand-Maillet, and Pun]{muller2001performance}
Henning M{\"u}ller, Wolfgang M{\"u}ller, David~McG Squire, St{\'e}phane
  Marchand-Maillet, and Thierry Pun.
\newblock Performance evaluation in content-based image retrieval: overview and
  proposals.
\newblock \emph{Pattern recognition letters}, 22\penalty0 (5):\penalty0
  593--601, 2001.

\bibitem[Radford et~al.(2021)Radford, Kim, Hallacy, Ramesh, Goh, Agarwal,
  Sastry, Askell, Mishkin, Clark, Krueger, and
  Sutskever]{Radford_Kim_Hallacy_Ramesh_Goh_Agarwal_Sastry_Askell_Mishkin_Clark}
Alec Radford, Jong~Wook Kim, Chris Hallacy, Aditya Ramesh, Gabriel Goh,
  Sandhini Agarwal, Girish Sastry, Amanda Askell, Pamela Mishkin, Jack Clark,
  Gretchen Krueger, and Ilya Sutskever.
\newblock Learning transferable visual models from natural language
  supervision.
\newblock \penalty0 (arXiv:2103.00020), Feb 2021.
\newblock URL \url{http://arxiv.org/abs/2103.00020}.
\newblock arXiv:2103.00020 [cs].

\bibitem[Reimers \& Gurevych(2019{\natexlab{a}})Reimers and
  Gurevych]{Reimers_Gurevych_2019}
Nils Reimers and Iryna Gurevych.
\newblock Sentence-bert: Sentence embeddings using siamese bert-networks.
\newblock \penalty0 (arXiv:1908.10084), Aug 2019{\natexlab{a}}.
\newblock URL \url{http://arxiv.org/abs/1908.10084}.
\newblock arXiv:1908.10084 [cs].

\bibitem[Reimers \& Gurevych(2019{\natexlab{b}})Reimers and
  Gurevych]{sentence-transformers/msmarco-bert-base-dot-v5_Hugging_Face}
Nils Reimers and Iryna Gurevych.
\newblock Sentence-bert: Sentence embeddings using siamese bert-networks.
\newblock In \emph{Proceedings of the 2019 Conference on Empirical Methods in
  Natural Language Processing}. Association for Computational Linguistics, 11
  2019{\natexlab{b}}.
\newblock URL \url{http://arxiv.org/abs/1908.10084}.

\bibitem[Rethmeier \& Augenstein(2023)Rethmeier and
  Augenstein]{rethmeier2023primer}
Nils Rethmeier and Isabelle Augenstein.
\newblock A primer on contrastive pretraining in language processing: Methods,
  lessons learned, and perspectives.
\newblock \emph{ACM Computing Surveys}, 55\penalty0 (10):\penalty0 1--17, 2023.

\bibitem[Rogers(1963)]{Rogers_1963}
F.~B. Rogers.
\newblock Medical subject headings.
\newblock \emph{Bulletin of the Medical Library Association}, 51\penalty0
  (1):\penalty0 114–116, Jan 1963.
\newblock ISSN 0025-7338.

\bibitem[Schuhmann et~al.(2022)Schuhmann, Beaumont, Vencu, Gordon, Wightman,
  Cherti, Coombes, Katta, Mullis, Wortsman, Schramowski, Kundurthy, Crowson,
  Schmidt, Kaczmarczyk, and
  Jitsev]{laion/CLIP-convnext_large_d.laion2B-s26B-b102K-augreg_hugging_face}
Christoph Schuhmann, Romain Beaumont, Richard Vencu, Cade~W Gordon, Ross
  Wightman, Mehdi Cherti, Theo Coombes, Aarush Katta, Clayton Mullis, Mitchell
  Wortsman, Patrick Schramowski, Srivatsa~R Kundurthy, Katherine Crowson,
  Ludwig Schmidt, Robert Kaczmarczyk, and Jenia Jitsev.
\newblock {LAION}-5b: An open large-scale dataset for training next generation
  image-text models.
\newblock In \emph{Thirty-sixth Conference on Neural Information Processing
  Systems Datasets and Benchmarks Track}, 2022.
\newblock URL \url{https://openreview.net/forum?id=M3Y74vmsMcY}.

\bibitem[Stutz et~al.(2023)Stutz, Roy, Matejovicova, Strachan, Cemgil, and
  Doucet]{Stutz_2023}
David Stutz, Abhijit~Guha Roy, Tatiana Matejovicova, Patricia Strachan,
  Ali~Taylan Cemgil, and Arnaud Doucet.
\newblock Conformal prediction under ambiguous ground truth.
\newblock \penalty0 (arXiv:2307.09302), Jul 2023.
\newblock URL \url{http://arxiv.org/abs/2307.09302}.
\newblock arXiv:2307.09302 [cs, stat].

\bibitem[Van~Rossum \& Drake~Jr(1995)Van~Rossum and Drake~Jr]{van1995python}
Guido Van~Rossum and Fred~L Drake~Jr.
\newblock \emph{Python reference manual}.
\newblock Centrum voor Wiskunde en Informatica Amsterdam, 1995.

\bibitem[Vovk et~al.(2005)Vovk, Gammerman, and Shafer]{vovk2005algorithmic}
Vladimir Vovk, Alexander Gammerman, and Glenn Shafer.
\newblock \emph{Algorithmic learning in a random world}, volume~29.
\newblock Springer, 2005.

\bibitem[Yang et~al.(2023)Yang, Shi, Wei, Liu, Zhao, Ke, Pfister, and
  Ni]{yang2023medmnist}
Jiancheng Yang, Rui Shi, Donglai Wei, Zequan Liu, Lin Zhao, Bilian Ke,
  Hanspeter Pfister, and Bingbing Ni.
\newblock Medmnist v2-a large-scale lightweight benchmark for 2d and 3d
  biomedical image classification.
\newblock \emph{Scientific Data}, 10\penalty0 (1):\penalty0 41, 2023.

\bibitem[Zhang et~al.(2023{\natexlab{a}})Zhang, Xu, Usuyama, Bagga, Tinn,
  Preston, Rao, Wei, Valluri, Wong, Lungren, Naumann, and
  Poon]{microsoft/BiomedCLIP-PubMedBERT_256-vit_base_patch16_224_Hugging_Face}
Sheng Zhang, Yanbo Xu, Naoto Usuyama, Jaspreet Bagga, Robert Tinn, Sam Preston,
  Rajesh Rao, Mu~Wei, Naveen Valluri, Cliff Wong, Matthew Lungren, Tristan
  Naumann, and Hoifung Poon.
\newblock Large-scale domain-specific pretraining for biomedical
  vision-language processing, 2023{\natexlab{a}}.
\newblock URL \url{https://arxiv.org/abs/2303.00915}.

\bibitem[Zhang et~al.(2023{\natexlab{b}})Zhang, Xu, Usuyama, Bagga, Tinn,
  Preston, Rao, Wei, Valluri, Wong, Lungren, Naumann, and
  Poon]{Zhang_Xu_Usuyama_Bagga_Tinn_Preston_Rao_Wei_Valluri_Wong_et_al_2023}
Sheng Zhang, Yanbo Xu, Naoto Usuyama, Jaspreet Bagga, Robert Tinn, Sam Preston,
  Rajesh Rao, Mu~Wei, Naveen Valluri, Cliff Wong, Matthew~P. Lungren, Tristan
  Naumann, and Hoifung Poon.
\newblock Large-scale domain-specific pretraining for biomedical
  vision-language processing.
\newblock \penalty0 (arXiv:2303.00915), Mar 2023{\natexlab{b}}.
\newblock \doi{10.48550/arXiv.2303.00915}.
\newblock URL \url{http://arxiv.org/abs/2303.00915}.
\newblock arXiv:2303.00915 [cs].

\end{thebibliography}
\clearpage

\appendix
\section{Appendix}

    \subsection{Overview}
        \label{app:overview}
        \begin{figure}[!h]
          \centering
          \includegraphics[width = 0.9\textwidth]{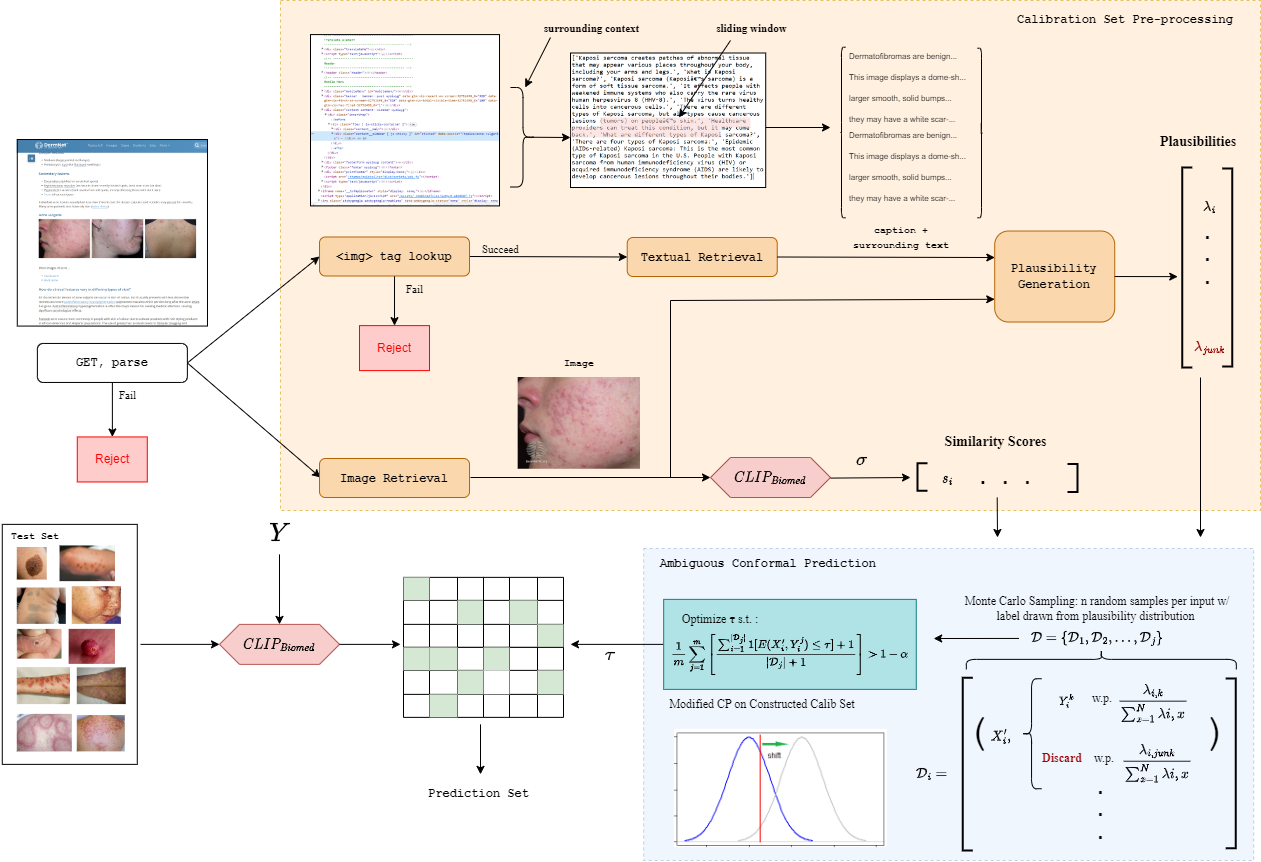}
          \caption{Figure portraying the total pipeline of data mining, plausibility generation, and conformal prediction}
        \end{figure}
        \begin{figure}[!h]
          \centering
          \includegraphics[width = 0.9\textwidth]{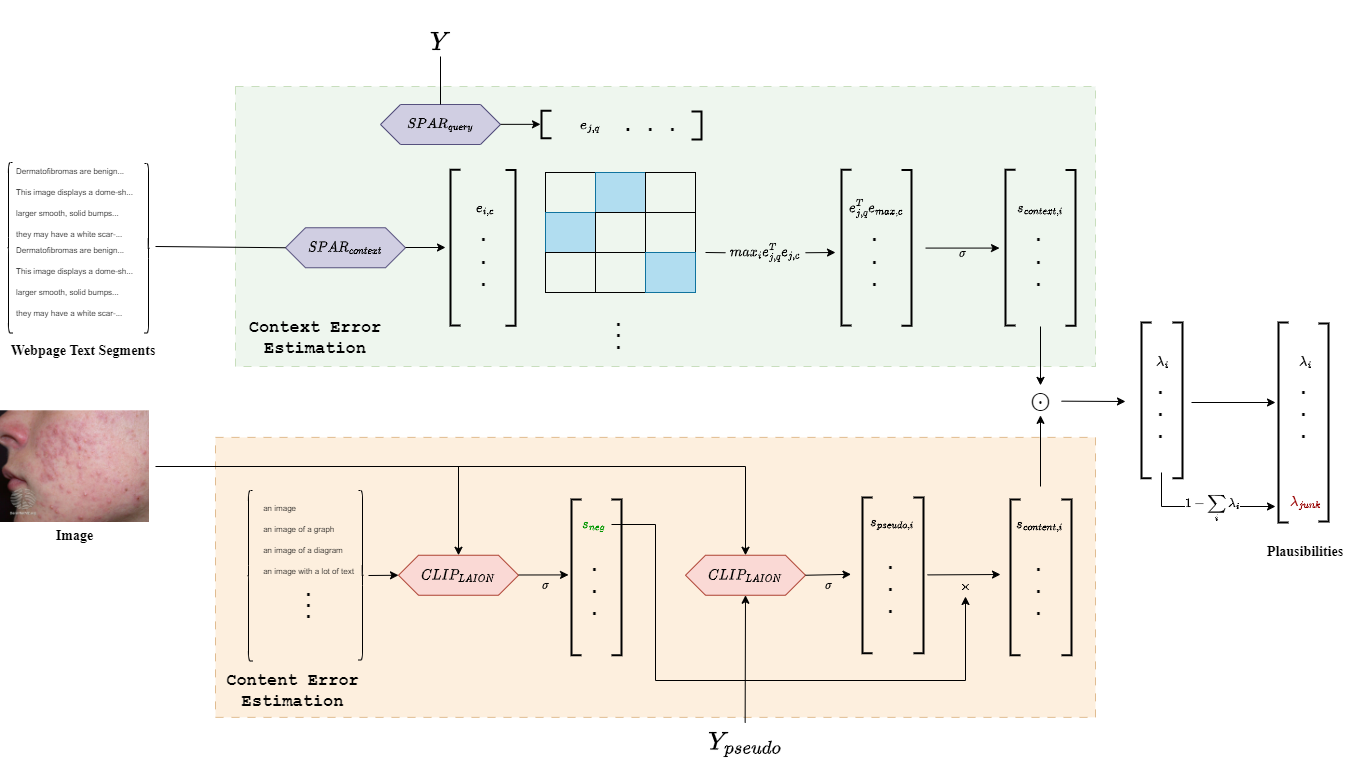}
          \caption{Figure portraying the hybrid context and content alignment based plausibility generation pipeline.}
        \end{figure}
        \pagebreak
        
    \subsection{Data Mining}
        \label{app:data_mining}
        \begin{algorithm}[!h]
            \caption{Web Scraping Procedure for Acquiring Internet Meta-Data ($\mathcal{C}_{web}$)}
            \label{alg:web_scraping}
            \textbf{Given:} A set of categories $\mathcal{Y}$, an image search engine $\mathcal{E}$ providing entries consisting of images, a corresponding URL to the image, and a corresponding URL to the image's web page source HTML; a target number of results per class, $K$.
            \begin{enumerate}
                \item For each category $y \in \mathcal{Y}$:
                \begin{enumerate}
                    \item Perform a query on $\mathcal{E}$ for $y$ to obtain a list of entries $\mathcal{L}_y$ with length $>> K$. 
                    \item Until the target number of examples $K$ is achieved, consider each entry in $\mathcal{L}$ consisting of an image ({\color{purple}{\texttt{image}}}), its URL ({\color{purple}{\texttt{image\_url}}}) and a URL to its web page source ({\color{purple}{\texttt{context\_url}}}). We find the location of the image in the web page linked to {\color{purple}{\texttt{context\_url}}} by identifying the presence of the string {\color{purple}\texttt{image\_url}} in that web page, and obtain the contextual meta-data immediately surrounding that location. Particularly:
                    \begin{enumerate}
                        \item If {\color{purple}{\texttt{context\_url}}} is unaccessible (e.g. timeout, blocked, or lazy loading) or there is no close match for the {\color{purple}\texttt{image\_url}} in the {\color{purple}\texttt{src}} or {\color{purple}\texttt{url-src}} tag of any {\color{purple}\texttt{<img>}} div in the webpage, skip and continue. We define a close match for an {\color{purple}\texttt{image\_url}} to be a {\color{purple}\texttt{src}} or {\color{purple}\texttt{url-src}} whose file-name (i.e. without file paths or arguments passed in with the URL) roughly matches the file-name in {\color{purple}\texttt{url-src}} ($>.85$ similarity, according to a metric on their similarity utilizing the \texttt{difflib.SequenceMatcher} library in Python ( \cite{van1995python}).
                        \item Otherwise, we retrieve the {\color{purple}\texttt{alt}} tag for the matching {\color{purple}\texttt{<img>}} divider in the context HTML page, and obtain its plaintext (if it is in HTML format). We also retrieve the minimum of 256 plaintext tokens or 10 sentences (ending each divider as a separate sentence) from the text immediately before the pertinent {\color{purple}\texttt{<img>}} tag, and from the text immediately after the tag, performing sentencing using the natural language tooklit in Python (\cite{Bird_Klein_Loper_2009}). We concatenate these surrounding plaintext results together.
                        \item We take {\color{purple}\texttt{image}} and all its concatenated surrounding plaintext sentences, and add these to $\mathcal{C}_{web}$ as an entry.
                    \end{enumerate}
                \end{enumerate}
            \end{enumerate}
            \textbf{Output:} a set of $K\cdot\#(\mathcal{Y})$ results for each of the $\#(\mathcal{Y})$ categories, with each result containing the image and its surrounding captions/images.
        \end{algorithm}
        \pagebreak

    \subsection{Experiments}
        \begin{figure}[!h]
            \centering
            \includegraphics[width = 0.7\textwidth]{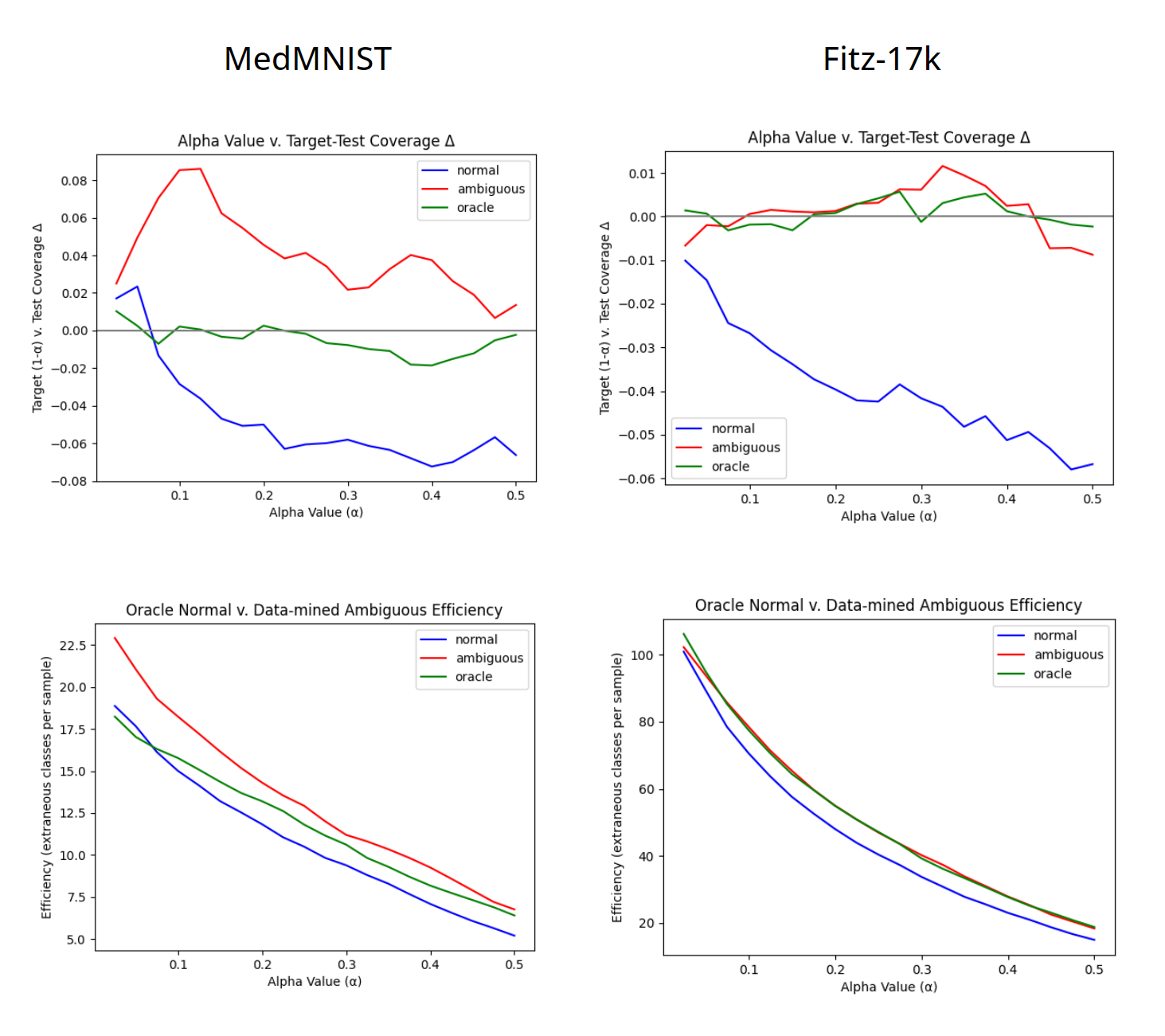}
            \caption{$\Delta_{coverage}$ and Efficiency for normal, oracle, and ambiguous CP across varying $\alpha$ values}
        \end{figure}

        \begin{figure}
          \centering
          \includegraphics[width = 0.9\textwidth]{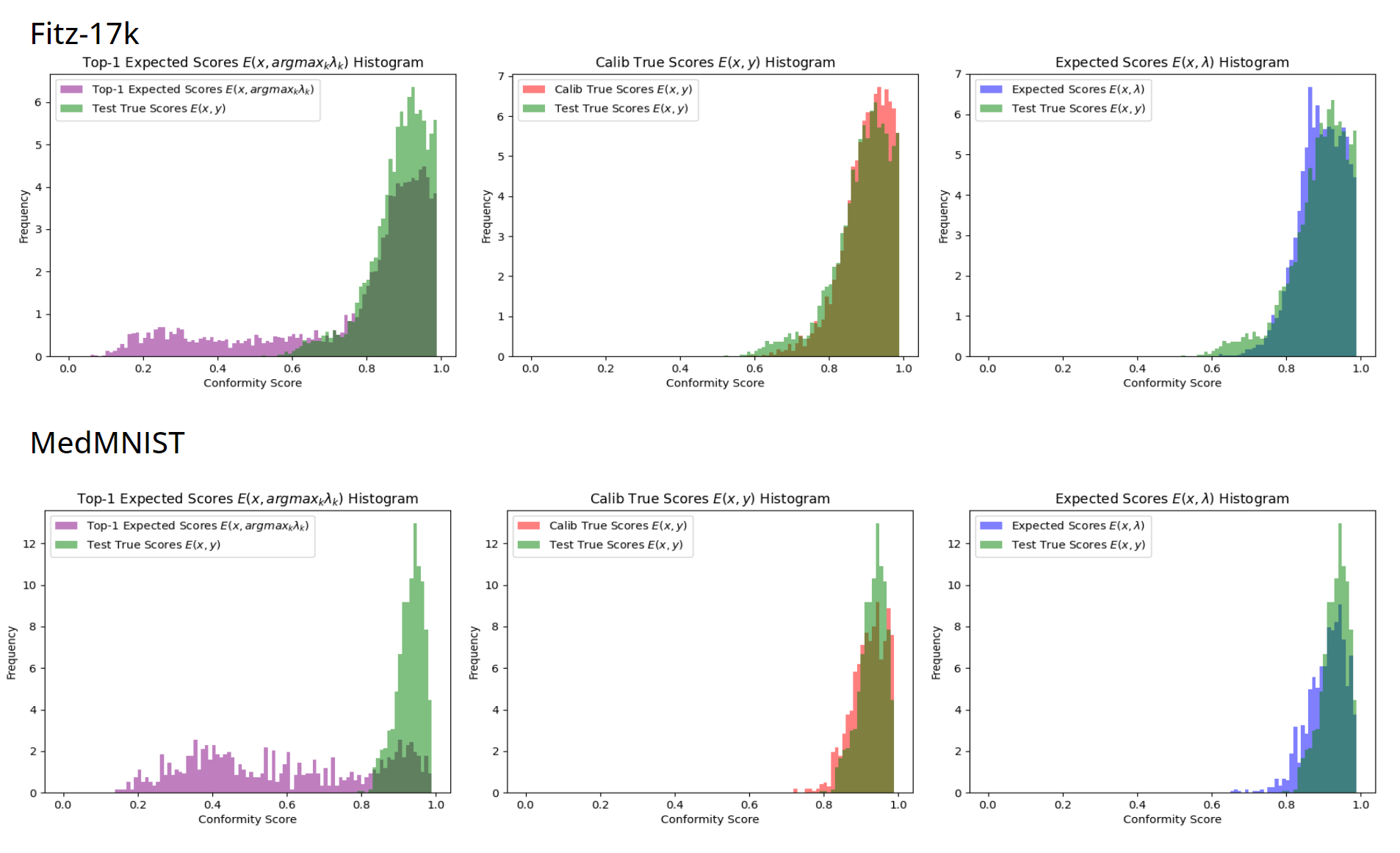}
          \caption{Conformity score distribution for FitzPatrick-17k and MedMNIST}
        \end{figure}

\end{document}